\documentclass[titlepage,oneside,12pt]{article}
\usepackage{lineno}
\usepackage{float}
\usepackage[tiny,rm]{titlesec}
\newpagestyle{trbstyle}{
\sethead{Aliari, F. Sadabadi}{}{\thepage}
}
\titleformat{\section}{\bfseries}{}{0pt}{\uppercase}
\titlespacing*{\section}{0pt}{12pt}{*0}
\titleformat{\subsection}{\bfseries}{}{0pt}{}
\titlespacing*{\subsection}{0pt}{12pt}{*0}
\titleformat{\subsubsection}{\itshape}{}{0pt}{}
\titlespacing*{\subsubsection}{0pt}{12pt}{*0}

\usepackage{enumitem}
\setlist[1]{labelindent=0.5in,leftmargin=*}
\setlist[2]{labelindent=0in,leftmargin=*}

\usepackage{ccaption}
\usepackage{amsmath}
\makeatletter
\renewcommand{\fnum@figure}{\textbf{FIGURE~\thefigure} }
\renewcommand{\fnum@table}{\textbf{TABLE~\thetable} }
\makeatother
\captiontitlefont{\bfseries \boldmath}
\captiondelim{\;}

\usepackage{times}

\usepackage[T1]{fontenc}
\usepackage{textcomp}

\usepackage[sort,numbers]{natbib}

\setcitestyle{round}

\usepackage{graphicx}
\usepackage{multirow}

\begin{document}

\title{Automatic Detection of Major Freeway Congestion Events Using Wireless Traffic Sensor Data: A Machine Learning Approach}
\author{
\textbf{Sanaz Aliari } (corresponding author) \\
Graduate Research Assistant \\
Department of Civil \& Environmental Engineering \\
University of Maryland  \\
College Park, MD 20740  \\
Email: saliyari@umd.edu\\ 
\vspace{10pt}\\
\textbf{Kaveh F. Sadabadi} \\
Senior Faculty Specialist \\
Center for Advanced Transportation Technology \\
University of Maryland \\
College Park, MD 20740  \\
Email: kfarokhi@umd.edu
\vspace{10pt}\\
Word count: 4,559 words text + (2 tables) x 250 words (each) = 5,059 words}

\thispagestyle{empty}
\maketitle
\setpagewiselinenumbers

\newpage

\thispagestyle{empty}
\section{Abstract}

Monitoring the dynamics of traffic in major corridors can provide invaluable insight for traffic planning purposes. An important requirement for this monitoring is the availability of methods to automatically detect major traffic events and to annotate the abundance of travel data. This paper introduces a machine learning based approach for reliable detection and characterization of highway traffic congestion events from hundreds of hours of traffic speed data.
Indeed, the proposed approach is a generic approach for detection of changes in any given time series, which is the wireless traffic sensor data in the present study. The speed data is initially time-windowed by a ten-hour long sliding window and fed into three Neural Networks that are used to detect the existence and duration of congestion events (slowdowns) in each window. The sliding window captures each slowdown event multiple times and results in increased confidence in congestion detection. The training and parameter tuning are performed on 17,483 hours of data that includes 168 slowdown events. This data is collected and labeled as part of the ongoing probe data validation studies at the Center for Advanced Transportation Technologies (CATT) at the University of Maryland. The Neural networks are carefully trained to reduce the chances of over-fitting to the training data. The experimental results show that this approach is able to successfully detect most of the congestion events, while significantly outperforming a heuristic rule-based approach. Moreover, the proposed approach is shown to be more accurate in estimation of the start-time and end-time of the congestion events.
\newline
\newline \textit{Keywords}: Congestion Detection, Slowdown Analysis, Machine Learning, Neural Networks, Time Series Change Detection, Anomaly Detection, Automatic Annotation, Wireless re-identification traffic monitoring (WRTM) technology
\newpage

\section{Introduction}
This paper deals with the problem of automatic annotation of congestion events on freeways using travel time data. The proposed approach is based on machine learning and results in an estimate of the start and end times of each event. Spotting congestion events on the huge network of roadways is a challenging task to be done manually. 
\par
The excessive population growth and development in urban areas resulted in the continuous increase in the number of vehicles on the road network. Monitoring the dynamics of traffic in major corridors can provide valuable information on traffic disruptions such as major slowdowns and bottlenecks. Detecting and reporting traffic anomalies in a timely manner provides additional insights to authorities for planning purposes such that the planned solutions could handle the future increases in the traffic load more effectively.
\par
Another application of slowdown detection is to validate probe speed data by comparing it with ground-truth data collected by Wireless Re-identification Traffic Monitoring (WRTM) technology. WRTM technology is a traffic data consolidating method based on signal detection of on-board devices such as Bluetooth, Wi-Fi and other wireless traffic monitoring devices. WRTM makes it possible to anonymously track a vehicle for the traffic monitoring purposes, by detecting the signals emitted from electronic devices in the vehicle to retrieve the travel time and speed data for a specific road segment. As such, WRTM leverages the abundance of data available on the behavior of users in the physical world due to emergence of Internet of Things (IoT) and increased number of users and provides data that can be used as ground truth to validate the quality and accuracy of GPS probe travel time data. Vehicle Probe Project (VPP) founded by I-95 corridor coalition, consolidates real time travel information from both freeways and major arterials using the WRTM technology ~\cite{Haghani2010,Haghani2009} to validate the GPS probe data, in terms of its punctuality and accuracy on capturing congestion events. The punctuality is validated based on latency analysis of the probe data, which measures the difference between the time the traffic flow is perturbed and the time that the change in speed is reflected in the probe data~\cite{Wang2017}. To that aim, precise detection of on-sets and off-sets of traffic perturbation events (congestions) are required. The accuracy of the probe data is examined based on slowdown analysis which measures the accuracy of probe data to capture the magnitude and duration of slowdowns. 

\par
\begin{figure}
\begin{center}\includegraphics[width=4.5in]{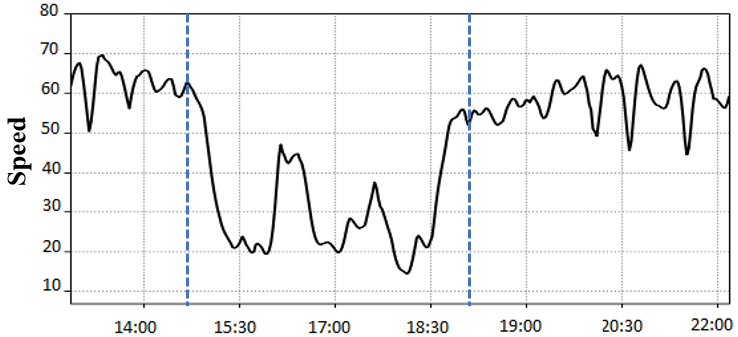}\end{center}
\caption{A visualization of speed data; slowdown period is marked with vertical lines.}
\label{fig:bec1}
\end{figure}

The focus of this paper is the detection of slowdown events on the targeted area of a freeway, as the first step in latency and slowdown analysis of the probe data using machine learning methods. In this study, a significant perturbation in traffic is considered a major slowdown when traffic speed reduces by at least 15 mph for a period of one hour or more~\cite{Sharifi2016}. An example of such event is depicted in Figure~\ref{fig:bec1}. The proposed approach is a two-stage machine learning based approach that leverages the data collected and labeled during the course of VPP to provide a reliable algorithm for detection of slowdowns. The core of the algorithm is the use of three independent Neural Networks that are trained to detect congestion event and to estimate the start-time and the end-time of slowdown events on any given ten-hour long window of speed data.

\par
We show that our approach can be significantly more reliable than a rule-based approach in detection of congestion events, and that it can perform well in case of new (unseen) data. The training and parameter tuning are performed on 12,443 hours of data and then tested on 5,040 hours of unseen speed data. It is shown that for training and validation datasets the slowdown detection Neural Network achieves a 98\% accuracy in detecting windows of speed data that contain a slowdown event. Also, the Neural Networks that are used for estimation of start-time and end-time of slowdown events, achieve less than 5\% mean squared estimation error for both training and validation datasets. The overall two-stage slowdown detection algorithm can successfully detect 87.5\% of the slowdowns in the unseen data, which shows that the proposed approach generalizes reasonably well to unseen data.

The rest of the paper is organized as follows. First a brief review of the existing slowdown detection methods is presented. Then the methodology of the machine learning-based slowdown detection is discussed. A detailed description of the dataset and the architecture of the Neural Networks used in this approach are then presented. subsequently, a comparison between the machine learning and rule-based approach is conducted in the experimental results section. Finally, key findings and direction for future work are summarized.

\section{Literature review}
Reliable detection of slowdowns is a challenging task as it deals with hundreds of hours of noisy speed data, which makes it a tedious task to be performed manually. Several heuristic rule-based approaches are suggested in the literature for automatic detection of congestion events in probe traffic speed data. A rule-based pattern recognition approach is introduced in~\cite{Wang2018}, in which slowdowns are detected as a period of time that recorded speeds fall below a percentage of free flow speed. The initial detections are then refined by few heuristic rules. In~\cite{Adu2015} a pattern recognition approach based on Empirical Mode Decomposition (EMD) is introduced that decomposes probe-sourced traffic speed data into distinct time-frequency speed trends, which present various time resolutions of speed oscillations representing short-term, mid-term and long-term trends of the speed data. The variations of these three are subsequently used to detect slowdown events at different time scales. 
In~\cite{ALTINTASI2017} a search algorithm is used to identify traffic congestion patterns in arterial roads using floating car data. Average travel speeds are translated into four qualitative state of traffic according to the Level of Service for arterials. Different traffic patterns are defined based on all possible combinations of the states.

Rule-based approaches may work well on the data used for their development. However, they usually do not perform as well on new data without re-tuning of their parameters.
Given the very large size of the speed data, parameter re-tuning is not usually convenient. Moreover, parameter tuning for one dataset, may damage the performance for other sets formerly used for development of the data. Indeed, the highly variable dynamics of the speed data, makes it difficult to have rules that apply to the entire dataset. 
In other words, the ability of the rule-based methods in providing a generic solution for datasets with different underlying dynamics can be inherently limited. Also, it is argued in~\cite{Wei2018} that rule-based methods that rely on a prior parametric model on the signal (utilizing features such as mean, variance and spectrum) can only detect statistically detectable boundaries and fail to detect more complex changes in dynamics of a time series. As a result, a semi-automatic approach is often employed: first a rule-based algorithm is used to create an initial list of candidate slowdowns and then, the candidate list is checked manually by an expert to ensure they actually correspond to slowdowns. Usually it happens that a large part of the slowdowns is missed by the rule-based algorithm and the manual screening is practically performed over the whole new dataset. 

These limitations, motivates the pursue of a Machine Learning(ML) approach that can leverage the abundance of formerly labeled data and provide a solution that can be generalized to new data, and even use the new data to improve its robustness. Artificial Neural Networks are one of the most popular ML approaches that are shown to have the ability to detect complex patterns from extremely noisy data. When provided with enough labeled training data, they are shown to be more robust than classical approaches in various disciplines of science. When trained properly to avoid over-fitting to the training dataset, they are also shown to be able to reasonably generalize well to unseen data~\cite{dropout2014}. 

If we regard the problem of slowdown detection in the speed data as the general problem of detecting anomalies (or changes) in a time-series, a number of related recent approaches can be found that employ machine learning techniques to detect special events in a given time series. In~\cite{Wei2018}, a deep learning approach is introduced based on the concept of Autoencoders that can automatically extract features specific to the data, without making any prior assumptions on the underlying generative processes. The approach in~\cite{Wei2018}, includes the use of a set of Autoencoders that are trained to ``encode'' multiple channels of the input data (from different sensors in an IoT system) into a set of features that capture the important dynamics of the input data. The distance between features extracted consecutive frames of inputs data is computed, and the points in time where this distance becomes higher than a threshold are classified as change-points. Another popular machine learning approach for change detection is the Long Short Term Memory (LSTM) networks that are able to learn long term patterns of unknown length in sequences, due to their ability to maintain long term memory~\cite{LSTM2015}. A stacked LSTM network approach for anomaly detection is introduced in~\cite{LSTM2015}, in which the network is trained on non-anomalous data and used as a predictor over time. Subsequently, the prediction errors are modeled as a multivariate Gaussian distribution, which is used to assess the likelihood of anomalous behavior. 
The proposed approach in this study, employs a two-stage window-based approach that is specifically designed to reduce sparsity of slowdown events in the training data and also to provide higher confidence in the final slowdown detection.  

\section{Methodology}
\begin{figure}
\begin{center}\includegraphics[width=5.5in]{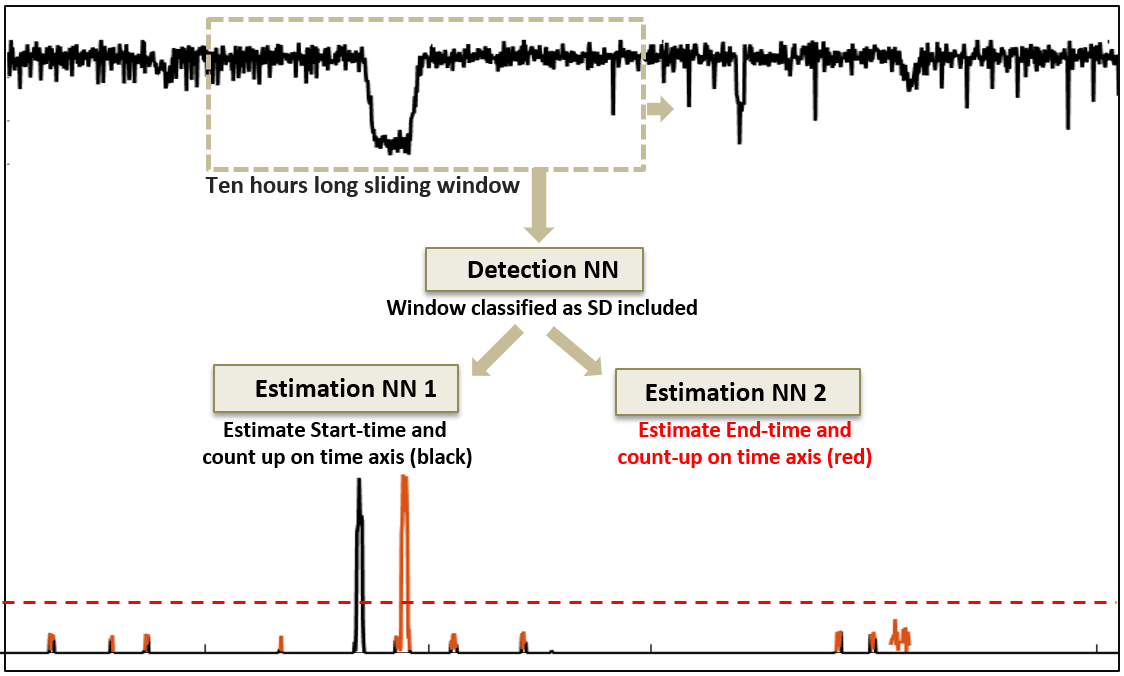}\end{center}
\vspace{-10pt}
\caption{The high level architecture of the proposed solution.}
\label{fig:hla2}
\end{figure}

In this study, a novel two-stage windowing approach based on Neural Networks(NN) is introduced for detection of slowdowns(SD). The high-level architecture of the solution is shown in Figure~\ref{fig:hla2}. In the first stage, a sliding window is moved over the speed data and for each window, a first NN(Detection NN) is used to classify the window as either a `SD-included' or a `non SD-included' window. If the window was classified as `SD-included', two other NNs are used to estimate the start-time and end-time of the detected SD (relative to the start time of the current window). As the window slides, a single SD  event would be detected multiple times, but the absolute start/end times are expected to match. As such, in the first stage, the number of start/end times detected at any given point are counted. 
In the second stage, the final list of candidate SDs is generated by including the points in time where the number of detected start/end-times are significantly high. As such, the use of the sliding window helps us to achieve higher confidence in SD detection: a real slowdown is expected to be detected tens of times, whereas erroneous detections are expected to be scattered on the time axis (hence we can easily remove them by a loose threshold).

Another important advantage of using an sliding window, is to overcome the issue of rare occurrence of SD events. The dataset we used for the current study, contains a total 17,483 hours of speed data in which, only 168 labeled SD events exist. Considering the one-minute time resolution of data, this means that only less than 0.001\% of the data points are labeled as SD events (start-time or end-time of a SD), and the remaining data points are labeled as non-SD. Training of an ML based approach using imbalanced dataset is known to introduce severely negative impact in the overall performance~\cite{Imbalance2015}. Data augmentation is a common techniques to circumvent the under-representation issue in machine learning problems, especially in the field of computer vision, where perturbations of an image that leave the underlying class unchanged are used to generate additional examples of the under-represented class~\cite{Augm2014}. Examples of such  perturbations include different clippings of a large image. The sliding window, plays a similar role during the training of our approach, as its continuous movement on the speed data (with one minute sliding steps) creates hundreds of examples of new SD events out of every single SD event in the database. Since the length of the sliding window is set to ten hours, it may happen that more than one major slowdowns occur in a window. For the training of NNs, such windows are labeled with reference to the {\it first} slowdown that is completely captured in the window (from start time to the end time). The NNs are expected to learn this convention and behave accordingly on new data.    

\subsection{Data Description}
In this study, the synthesized Bluetooth and WiFi WRTM data is used to train and test the Neural Networks. The WRTM data collection is based on deployment of portable sensors within known and pre-specified distances of road segments and matching the detected Media Access Control addresses (MAC address) between two sensors to estimate the travel time and speed. A MAC address is a hardware identification number that uniquely identifies each device on a network. Traffic Message Channel (TMC) is defined and standardized by Traveller Information Services Association (TISA) international organization, for location referencing~\cite{tisa}.  Data collection segments may include one or more Traffic Message Channel (TMC) base segments, such that the total length of each data
collection segment is usually one mile or greater for freeways. Speed data is sampled at one-minute time intervals. The details of the speed data preparation is presented in~\cite{Wang2017}. The data used for training and testing includes 17,483 hours of data collected in states of Pennsylvania, Georgia, New Hampshire and South Carolina. This data includes 168 slowdowns, spanning a total of 538 hours. An overview of data collection effort is presented in Table \ref{table:data1}.

\subsection{Neural Networks Architecture}
The detection NN is used to classify any window of speed data into one of the two classes: `SD-included' and `non SD-included'. Since the length of SD events included in the dataset are between one hour to six hours, the length of window is set to ten hours (600 minutes). Given the one minute resolution of the speed data this means that the input to the NN is a vector of 600 speed values. 
We use a two-layer Fully Connected Neural Network (FCNN) that consists of 70 neurons in each layer. The structural details of the FCNN can be found in~\cite{Zach2017}. Since the detection NN is a classifier, a cross-entropy loss function~\cite{lossEntropy2005} is used as an appropriate choice for this NN. This means that the output of the detection NN is initially a two-element vector containing the probabilities of each window belonging to any one of the two classes (the sum of values should be equal to one). Cross-entropy minimization of two Probability Distribution Functions (PDF) can be achieved when the two PDFs (labels and NN outputs in our case) are matched. 

We use Rectified Linear Units (ReLU) as the activation function for the inner layer as they are  tend to show better convergence performance during training, when back propagation approach is used for training~\cite{Relu2012} (they are less prone to the vanishing gradients issue in back-propagation). However, a sigmoid activation function is used for the output layer as it normalizes an input real value into the range between zero and one.

For the two estimation NNs we use two separate two-layers FCNNs, with 70 neurons at each layer.  
As these NNs are used for estimation of start-time and end-time of the windows containing an SD event, we use a Mean Squared Error (MSE) loss function for these two NN. Again, a Rectified Linear Units (ReLU) as the activation function for the inner layer and a sigmoid function for the output layer are used. 

\section{Experimental Results}
We used the speed data from states of Pennsylvania, Georgia and New Hampshire for training of the three Neural Networks. The South Carolina state data was put aside for final testing of the performance of the overall algorithm on unseen data. Applying the sliding-window on these three sets of data, a total of 65000 windows of speed data was created, in which 32000 were labeled as SD-included and 33000 were labeled as non SD-included. We use 80\% of these windows for training and the remaining 20\% was used for validation during training to make sure that the NNs are able to generalize to new unseen data. Moreover, 15\% of weights are randomly dropped-out of each training iteration to avoid over-fitting of the weights. Dropping out a random subset of weights at each training iteration, is shown to be effective against over-fitting of the Neural Networks to the training data~\cite{dropout2014}.

After training the Detection NN for 1000 epochs with a batch size 512, the training set classification accuracy reached 98.3\% and the validation set classification accuracy reached 98.18\%. With a similar training strategy, the start-time estimation NN achieved 4.88\% mean squared error rate for the training set and 4.84\% for the validation set. The end-time estimation NN achieved 4.84\% mean squared error rate for the training set and 5.00\% for the validation set. 

For evaluation of the performance of the overall two-staged solution, we use the following set of performance metrics:

\begin{itemize}[noitemsep]
\item True Positives (TP): the number of actual slowdowns that are correctly detected. 
\item False Negatives (FN): the number of slowdowns that are not detected. 
\item False Positives (FP): the number of detections not corresponding to an actual slowdown.
\end{itemize}
For computation of these measures of performance, a slowdown is considered to be correctly detected, if its start/end time were detected within a specified time margin. These performance measures are used to compare the performance of the proposed Machine Learning (ML) based solution  to the Rule-Based (RB) approach introduced in~\citep{Wang2018}. The results are summarized in Table~\ref{table:30min2} for a 30 minutes time margin (tolerance). Note that the South Carolina state data was not used for training of the ML based approach. It can be seen that the proposed approach significantly outperforms the rule-based approach, both in terms of True Positives and False Negatives. Not only this is true for the data used for training of the ML based approach, but also is true for the case of unseen South Carolina state data. This shows that the approach is not over-fit to the training dataset and can generalize reasonably well to new datasets.

\begin{figure}[h!]
\begin{center}\includegraphics[width=6.5in]{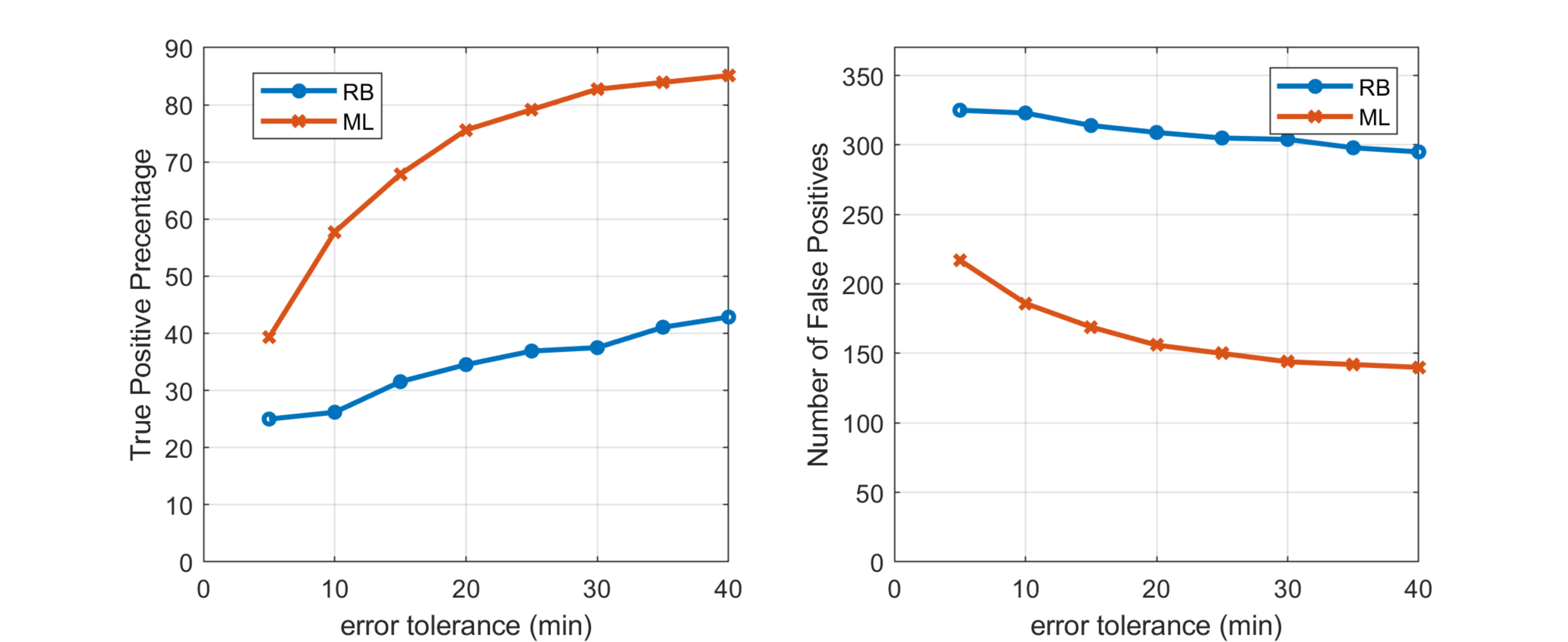}\end{center}
\caption{Percentage of True Positives (PTP) and the number of False Negatives versus different tolerances of error.  }
\label{fig:cmp3}
\end{figure}

It is also important to evaluate the accuracy of estimating the time of occurrence (start/end times) of slowdowns. To evaluate the estimation accuracy, we measured the variations of performance measures for different tolerances of error. Figure~\ref{fig:cmp3}, shows how the Percentage of True Positives (PTP), and the number of False Negatives vary versus the amount of error tolerances. Entire 17,483 hours of data consisting of 168 slowdown events is used for this experiment. It can be seen that, the proposed ML based approach achieves significantly higher PTPs even for smaller tolerances of error. For instance for the case of 10 minutes tolerance the PTP of the ML based approach is twice more than that of the Rule-Based approach. Similar observation can be made for the Number of False Positives. 

\section{Conclusions}
In this paper, a novel machine learning based approach is introduced for automatic detection of congestion events (Slowdowns) from the abundance of speed data. The proposed two-staged approach consists of using three independent Neural Networks responsible for tasks of detection of slowdowns,  estimation of start-time and estimation of end-time of each slowdowns. It is shown that the  proposed approach significantly outperforms a conventional rule-based approach, both in terms of percentage of true positive detections and number of false negatives. It was also shown that the  approach can generalize well to the case of unseen data. 

The results of this study are indeed indicative of the potential of machine learning in providing reliable tools for traffic monitoring and planning purposes. Apart from real-time monitoring of the network traffic, such a reliable approach for slowdown detection can be used to annotate the abundance of data collected every day on huge networks of highways, in order to create databases for more in-depth analysis and understanding of the inter-related traffic trends on different parts of the network.

The improvement in performance of the ML based approach can be a subject of future work. This may include evaluation of more sophisticated training approaches or the application of other well-known Neural Network architectures, such as convolutional neural networks that are shown to be powerful in recognizing complex patterns in the field of computer vision. 
Also, the future work may include the application of the proposed approach inside the latency analysis framework for validation of GPS probe data. One possibility is to apply the proposed algorithm to both GPS probe data and WRTM data, to extract and compare the corresponding start-times and end-times of slowdowns. The other possibility is to design a unified approach that takes both GPS and WRTM data and estimates the latency between them.
 
\section{Contribution Statement}

The authors confirm contribution to the paper as follows: study conception and design: S. Aliari, K. F. Sadabadi; data collection: Center for Advanced Transportation Technology (CATT), University of Maryland; analysis and interpretation of results: S. Aliari; draft manuscript preparation: S. Aliari. All authors reviewed the results and approved the final version of the manuscript.

\bibliographystyle{trb}
\bibliography{ref}
\newpage

\begin{table}[h!]
\centering
\vspace{40pt}
\caption{WRTM data collection summary}
\begin{tabular}{||l c c c c c c c||} 
 \hline
 &   &  &  & Directional & Num. of & Data Collection & Num. of   \\ State &  Road & Start & End& Length(mi)& Segments &Duration (hrs) & SD's \\[0.5ex] 
 \hline\hline
Pennsylvania & 	I-79	&  Exit 73 &	Exit 88  & 30 & 14 & 4,704 & 52 \\
\hline
Georgia &  	I-75	&  Exit 187 &	Exit 222  & 36 & 13 & 3,419 & 18 \\
\hline
\multirow{2}{*}{South Carolina}  & 	I-85	&  Exit 48 &	Exit 56  & 7.15 & 8 &\multirow{2}{*}{5,040}& \multirow{2}{*}{72} \\ & I-26 & Exit 108 & Exit 103 & 4.5 & 6 & &\\
 \hline
 \multirow{3}{*}{New Hampshire}  & 	I-89	&  Exit 5B &	Exit 3  & 2.2 & 2&\multirow{3}{*}{4,320} & \multirow{3}{*}{26} \\ &  I-93 & Exit 47 & Exit 44 & 2.6 & 2 &&\\&  I-93 & Exit 46A & Exit 37 & 14.5 & 8 &&\\
  \hline\hline
\end{tabular}

\label{table:data1}
\end{table}
\vspace{20pt}
\begin{table}[h!]
\centering
\caption{Slowdown detection comparison with 30 minutes tolerance. }
\begin{tabular}{||l c c c|c|c|c|c||} 
 \hline
State & Num. of segments & SD duration (hrs) & Num. of SDs & Method & TP & FN & FP\\[0.5ex] 
 \hline\hline
\multirow{2}{*}{Pennsylvania} &	\multirow{2}{*}{14}	&  \multirow{2}{*}{211} &	\multirow{2}{*}{52}  & ML & \textbf{42} & \textbf{10} & \textbf{20}\\\cline{5-8} &&&&  RB &6 & 46 & 59\\
\hline
\multirow{2}{*}{Georgia} &	\multirow{2}{*}{13}	&  \multirow{2}{*}{62.5} &	\multirow{2}{*}{18}  & ML & \textbf{15} & \textbf{3} & \textbf{7}\\\cline{5-8} &&&&  RB &9 & 9 & 33\\
\hline
\multirow{2}{*}{New Hampshire} &	\multirow{2}{*}{12}	&  \multirow{2}{*}{82.5} &	\multirow{2}{*}{26}  & ML & \textbf{23} & \textbf{3} & \textbf{48}\\\cline{5-8} &&&&  RB &9 & 17 & 71\\
 \hline
 \multirow{2}{*}{South Carolina} &	\multirow{2}{*}{14}	&  \multirow{2}{*}{182} &	\multirow{2}{*}{72}  & ML & \textbf{59} & \textbf{13} & \textbf{69}\\\cline{5-8} &&&&  RB &39 & 33 & 141\\
  \hline\hline
\end{tabular}
\label{table:30min2}
\end{table}


\end{document}